\documentclass[a4paper,fleqn]{cas-sc}
\usepackage[numbers]{natbib}
\usepackage{caption}
\usepackage{makecell} 
\usepackage{adjustbox} 
\usepackage{multirow} 
\usepackage{array} 
\usepackage{tabularx} 
\usepackage{booktabs} 
\usepackage{amsmath} 
\usepackage{amssymb}
\usepackage{graphicx}
\usepackage{subcaption}

\usepackage{lipsum}
\usepackage{hyperref}
\usepackage{flushend}
\usepackage{xcolor}
\usepackage{setspace}
\usepackage{float}
\usepackage{algorithm}
\usepackage{algpseudocode}

\usepackage[nomath]{lmodern}
\usepackage[T1]{fontenc}
\usepackage{fix-cm}           

\def\tsc#1{\csdef{#1}{\textsc{\lowercase{#1}}\xspace}}
\tsc{WGM}
\tsc{QE}

\begin{document}
\let\WriteBookmarks\relax
\def\floatpagepagefraction{1}
\def\textpagefraction{.001}
\let\printorcid\relax 

\shorttitle{PaAno+: Multiscale Encoding and Cross-Variable Attention for Time Series Anomaly Detection}

\shortauthors{Youji Zhu et al.}

\title[mode = title]{PaAno+: Multiscale Encoding and Cross-Variable Attention for Time Series Anomaly Detection} 

\author[1]{Youji Zhu}
\credit{Methodology, Software, Writing – original draft}

\author[2]{Hongbing Wang}
\credit{Supervision, Writing – review \& editing}

\author[1]{Wenchao Liu}
\credit{Visualization, Software}

\author[2]{Xiaodong Liu}
\credit{Visualization, Software}

\author[2]{Xiangguang Xiong\corref{cor1}}
\credit{Conceptualization, Methodology, Supervision, Writing – review \& editing}
\ead{xxg@gznu.edu.cn}

\address[1]{School of Mathematical Sciences, Guizhou Normal University, Guiyang 550025, China.} 
\address[2]{School of Big Data and Computer Science, Guizhou Normal University, Guiyang 550025, China.}

\cortext[cor1]{Corresponding author}  

\begin{abstract}
Time-series anomaly detection has significant practical value for industrial and medical monitoring, as well as other critical domains. Current Transformer- and large-model-based detection approaches incur excessive computational overhead, while existing lightweight alternatives are constrained by insufficient feature extraction and inadequate modeling of dependencies across multivariate variables. To mitigate the above drawbacks, this study develops a lightweight, efficient anomaly detection model, dubbed PaAno, within the patch-oriented representation learning paradigm. In the encoder module, a multiscale feature-extraction backbone is constructed using convolutional kernels with differentiated receptive fields to capture hierarchical temporal characteristics; subsequent cross-scale adaptive attention aggregation, combined with residual connection optimization, further stabilizes feature representation learning. A cross-variable fusion attention module is embedded to explicitly characterize inter-variable correlations, empowering the model to identify anomalous patterns amid intricate operational conditions. Moreover, a novel pretext task based on temporal patch-window sorting is customized to uncover intrinsic structural properties of time series, and triplet loss is leveraged to optimize the patch embedding space for enhanced feature discrimination. Extensive experiments on the TSB-AD benchmark demonstrate that the proposed PaAno achieves state-of-the-art detection accuracy on both univariate and multivariate tasks, yielding significant performance gains across evaluation metrics, including VUS-PR, relative to the original PaAno. Leveraging a compact network design, the presented model achieves favorable computational efficiency, enabling deployment on resource-limited terminals for real-time anomaly inference.
\end{abstract}

\begin{keywords}
Time-series anomaly detection \sep Patch representation learning \sep Multiscale convolution \sep Cross-variable attention \sep Self-supervised learning
\end{keywords}

\maketitle

\section{Introduction}
Time-series observations are chronologically ordered measurements widely deployed in industrial sensing, financial analytics, and clinical diagnosis \cite{Yue2022TS2Vec, Jia2024GPT4MTS}. Their inherent temporal dependence arises from contextual evolution and system dynamics \cite{Paparrizos2022VUS}. System malfunctions, external interference, or manual misoperation can disrupt this temporal correlation and cause anomalous behavior, including abrupt spikes, sharp declines, and persistent offsets. Time-series anomaly detection aims to detect abnormal samples or subsequences that deviate from the normal data distribution \cite{Wang2025ScaledBregman}. These irregularities generally indicate latent system risks; accordingly, accurate real-time detection is indispensable for reliable operational maintenance in practical applications \cite{Paparrizos2022VUS}. Given the latency and hardware constraints of edge terminals, deployed detection methods must balance prediction accuracy and inference efficiency within limited computing resources \cite{Wang2025ScaledBregman}.

Existing time-series anomaly detection techniques fall into three mainstream branches \cite{Liu2024Benchmark, Park2026PaAno}. The first branch covers statistical and conventional machine learning methods (Stat \& ML). Reconstruction-oriented approaches such as SubPCA \cite{Lu2004PCA} and RobustPCA \cite{Candes2011RPCA} quantify abnormality via the reconstruction residual after PCA-based dimensionality reduction. Cluster-driven methods, including KMeansAD \cite{Yairi2001FaultDetect}, KShapeAD \cite{Paparrizos2015kshape}, and CBLOF \cite{He2003ClusterOutlier}, partition normal samples into distinct clusters and calculate anomaly scores by inter-cluster distance metrics. Fitting-based models such as POLY \cite{Li2007StreamPoly} and SR \cite{Ren2019MSAD} derive residual values from polynomial or spectral fits to discriminate anomalies. Tree-based detectors, IForest \cite{Liu2008IF} and EIF \cite{Hariri2021EIF}, measure anomaly probability based on the path length of the sample separation. LOF \cite{Breunig2000LOF}, KNN \cite{Ramaswamy2000OutMine}, HBOS \cite{Goldstein2012HBOS}, and COPOD \cite{Li2020COPOD} identify outliers based on local density, neighbor spacing, or histogram-based statistics. MatrixProfile \cite{Yeh2016MatrixPro}, Series2Graph \cite{Boniol2020Series2Graph}, and SAND \cite{Boniol2021SAND} localize anomalies by computing subsequence similarity or constructing state-transition graphs. Free from the overhead of iterative training, these lightweight methods fit periodic, distribution-specified simple anomaly scenarios \cite{Liu2024Benchmark, Park2026PaAno}. Nevertheless, their fixed-distribution hypothesis impedes nonlinear local feature extraction; susceptibility to miscellaneous noise and neglect of sequential constraints degrade detection accuracy in the presence of complex nonlinear outliers \cite{Liu2024Benchmark}.

The second branch corresponds to deep neural network (NN)-based solutions. CNN \cite{Wang2017TSBaseline}, DeepAnT \cite{Munir2019DeepAnT}, and TimesNet \cite{Wu2023TimesNet} adopt 1D convolution to extract local temporal features and perform anomaly screening based on prediction or reconstruction residuals. LSTMAD \cite{Malhotra2015LSTMAD} leverages recurrent units to capture sequential dependence and identify anomalies via forecast bias. Classical autoencoder frameworks, including AutoEncoder \cite{Sakurada2014AEAD}, USAD \cite{Audibert2020USAD}, and OmniAnomaly \cite{Su2019SRNNAD}, adopt reconstruction error or adversarial optimization; specifically, OmniAnomaly integrates a stochastic recurrent structure and variational inference for probabilistic reconstruction. The frequency-domain method FITS \cite{Xu2024FITS} realizes compact network design via complex-domain interpolation. Compared with traditional statistics-based methods, deep learning architectures capture long-range temporal dependencies and achieve superior detection precision. Still, most existing networks fail to excavate cross-variable long-distance correlations. Excessive parameter count, sensitive window hyperparameter selection, and unstable convergence during training limit their performance on multivariate cross-period anomalous events \cite{Liu2024Benchmark, Park2026PaAno}.

The third branch consists of Transformer architectures and large pre-trained foundational models (Transformer). Self-attention-driven detectors, including AnomalyTransformer \cite{Xu2022AnomalyTrans}, TranAD \cite{Tuli2022TranAD}, DCdetector \cite{Yang2023DCDetector}, PatchTST \cite{Nie2023Word64}, and iTransformer \cite{Liu2024iTransformer}, identify anomalies through discrepancy calculation, adversarial optimization, contrastive attention learning, or block-wise prediction. Large pre-trained models MOMENT \cite{Goswami2024MOMENT}, TimesFM \cite{Das2024DecoderTS}, Chronos \cite{Ansari2024Chronos}, and LagLlama \cite{Rasul2023LagLlama} exploit masked modeling, decoder attention, temporal binning, and lag-covariate pre-training to support zero-shot or fine-tuning anomaly inference. Designed for global dependency modeling and cross-domain generalization enhancement, such paradigms exhibit promising application potential \cite{Liu2024Benchmark, Park2026PaAno}. However, billions of learnable parameters incur prohibitive memory and computational cost, hindering practical edge deployment and real-time online monitoring \cite{Park2026PaAno}. Recent empirical research indicates that biased evaluation protocols, such as inconsistent annotation rules and point-adjusted evaluation metrics, lead to inflated detection performance for large models and a misleading impression of technical advancement \cite{Liu2024Benchmark, Park2026PaAno}. Under standardized, unbiased evaluation rules, oversized foundation models achieve marginal performance gains over compact alternatives while retaining prohibitive computational costs \cite{Park2026PaAno}.

As a lightweight patch-driven representation-learning baseline, PaAno \cite{Park2026PaAno} achieves competitive detection accuracy but suffers from evident structural drawbacks. First, single-receptive-field fixed convolution kernels limit multiscale feature extraction, as they cannot simultaneously adapt to both pointwise abrupt anomalies and long-sequence continuous anomalies. Second, the model independently encodes each multivariate channel in separation mode, ignoring intrinsic inter-variable correlations and thereby weakening its ability to detect interaction-induced outliers. Third, its original auxiliary pretext task only constrains intra-patch temporal continuity, providing insufficient supervisory information for learning fine-grained temporal evolution and thereby limiting the depth of temporal dependency modeling. To remedy the aforementioned defects, this study proposes an improved lightweight detector named PaAno+, whose core contributions are summarized as follows:
\begin{itemize}
\item A multiscale temporal encoder is constructed with multi-branch parallel convolution, cross-scale adaptive attention aggregation, and residual shortcut connection for adaptive multigranularity feature extraction.
\item A cross-variable fusion attention module is proposed to execute multi-head self-attention along variable dimensions and explicitly model pairwise multivariate dependence.
\item A novel window-rearrangement pretext task shuffles patch windows and recovers the original temporal order, thereby generating adjustable regularization signals that strengthen the model’s awareness of temporal context variation.
\item Extensive validation on the public TSB-AD benchmark verifies that PaAno+ yields consistent accuracy gains in univariate and hybrid-type anomaly detection while preserving lightweight inference efficiency.
\end{itemize}

The rest of this paper is organized as follows. Section \ref{sec2} elaborates on related literature. Section \ref{sec3} details the structural design of PaAno+. Section \ref{sec4} presents the experimental setup and result analysis. Section \ref{sec5} concludes this work and outlines prospective research directions.

\section{Related work}
\subsection{Time Series Anomaly Detection}
\label{sec2}
Time-series anomaly detection aims to detect abnormal samples or subsequences that deviate from the normal data distribution \cite{Liu2024Benchmark, Park2026PaAno}. Based on label availability, existing solutions are grouped into supervised, semi-supervised, and unsupervised frameworks \cite{Darban2024Survey}. Due to the extreme sparsity of real-world anomaly annotations, semi-supervised methods trained exclusively on normal samples dominate in practical applications \cite{Park2026PaAno}. Classical statistical and machine learning detectors span multiple branches: density-driven LOF \cite{Breunig2000LOF}, boundary-constrained OCSVM \cite{Scholkopf2000SVM}, and isolation-based IForest \cite{Liu2008IF}; reconstruction-oriented PCA \cite{Lu2004PCA} and RobustPCA \cite{Candes2011RPCA}; and clustering-supported KMeansAD \cite{Yairi2001FaultDetect}, KShapeAD \cite{Paparrizos2015kshape}, and Series2Graph \cite{Boniol2020Series2Graph}. Such approaches offer concise implementation and clear interpretability but fail to capture the complex temporal dependencies embedded in raw data \cite{Liu2024Benchmark, Park2026PaAno}.

Deep learning has become a prevailing approach to time-series anomaly detection in recent years. Recurrent architectures such as LSTMAD \cite{Malhotra2015LSTMAD} and OmniAnomaly \cite{Su2019SRNNAD} explicitly model sequential dependencies and identify outliers via prediction or reconstruction residuals. CNN-based schemes, including DeepAnT \cite{Munir2019DeepAnT} and TimesNet \cite{Wu2023TimesNet}, extract local temporal patterns via convolutional operations. Transformer-type detectors, such as AnomalyTransformer \cite{Xu2022AnomalyTrans}, TranAD \cite{Tuli2022TranAD}, and PatchTST \cite{Nie2023Word64}, exploit self-attention to capture long-range temporal correlations. Furthermore, pre-trained time-series foundation models, including MOMENT \cite{Goswami2024MOMENT}, TimesFM \cite{Das2024DecoderTS}, and LagLlama \cite{Rasul2023LagLlama}, support anomaly inference via zero-shot prediction or parameter fine-tuning.

\subsection{Patch-Based Feature Learning}
Patch-oriented modeling has attracted growing research attention for contemporary time-series analytics. In time-series forecasting, PatchTST \cite{Nie2023Word64} integrates a Transformer with patch embedding, compressing the original sequence length \(L\) to \(L/p\) to reduce computational overhead and strengthen local pattern perception. TimesNet \cite{Wu2023TimesNet} reshapes 1D sequential input into 2D tensors and applies 2D convolution to capture intra-period and inter-period fluctuation characteristics simultaneously. Driven by such progress, patch modeling has been gradually extended to anomaly detection tasks. DCdetector \cite{Yang2023DCDetector} improves fine-grained anomaly discrimination through patch masking and dual-path contrastive learning, whereas the correlation difference mechanism of AnomalyTransformer \cite{Xu2022AnomalyTrans} relies primarily on local sliding windows and adheres to a patch-based modeling philosophy.

A series of dedicated patch-based anomaly detectors has been proposed. PaAno \cite{Park2026PaAno} adopts a lightweight 1D-CNN to project segmented time series into latent embeddings, optimizes the discriminative patch embedding space via triplet loss and an auxiliary classification pretext task, and implements online anomaly screening by matching test patches against pre-stored normal patch libraries; it achieves competitive detection accuracy with far fewer parameters than mainstream Transformer models. CATCH \cite{Wu2025CATCH} migrates the patch partition into the frequency domain and splits the spectrum into patches to retain high-frequency anomalous details. PatchAD \cite{Zhong2025PatchAD} substitutes the point-wise reconstruction loss with a patch-level objective by combining MLP-Mixer with masked reconstruction. Generally, patch-centric representation learning focuses on local temporal characteristics, reduces computational cost, and facilitates the detection of subtle outliers, serving as an essential technical path that balances lightweight network design and favorable detection performance \cite{Park2026PaAno, Xu2022AnomalyTrans}.

\subsection{Multiscale Features and Attention Mechanisms}
The joint use of multi-scale feature extraction and attention constitutes an effective modeling paradigm for temporal anomaly detection, simultaneously capturing fine-grained local details and global interdependencies. Three mainstream multi-scale modeling strategies have been developed. First, multi-sized parallel convolution kernels extract cross-scale local features, as in InceptionTime \cite{Fawaz2020InceptionTime} and MFAM-AD \cite{Xia2024MFAMAD}. Second, multi-resolution pyramid architecture builds hierarchical features via layer-wise downsampling, including Pyramid Transformer \cite{Zhang2024GM} and CSCAD \cite{Lee2025CSCAD}. Third, period-aware modeling, exemplified by TimesNet \cite{Wu2023TimesNet}, restructures 1D time series into periodic 2D tensors and leverages 2D convolution to extract intra-period features and cross-period variation rules.

Attention enhances feature modeling by separately modeling temporal and feature dimensions. Along the temporal dimension, long-range dependency can be well characterized: AnomalyTransformer \cite{Xu2022AnomalyTrans} distinguishes normal and abnormal data by the discrepancy between prior and sequential correlations. TranAD \cite{Tuli2022TranAD} captures global temporal trends via adversarial optimization. DCdetector \cite{Yang2023DCDetector} boosts detection precision via asymmetric dual-path attention. In the feature dimension, adaptive multi-scale feature fusion is achieved through attention weighting. MFAM-AD \cite{Xia2024MFAMAD} aggregates multi-granularity features with attention weights, and dual-memory architecture \cite{Gao2025VideoAD} embeds attention into memory-augmented modules to refine the final feature representation.

Existing multi-scale and attention fusion pipelines fall into two categories. The serial framework first performs multi-scale feature extraction and then further optimizes feature expression via attention; e.g., MFAM-AD \cite{Xia2024MFAMAD} adopts cross-scale attention to calibrate and align multi-level feature semantics. The parallel collaborative framework iteratively optimizes multi-scale construction and attention learning: the Pyramid Transformer \cite{Zhang2024GM} embeds downsampling and multi-task optimization within each pyramid layer, while the dual-memory model \cite{Gao2025VideoAD} establishes cross-scale information propagation across parallel branches. Multi-scale modeling addresses the representational deficiency of single-scale feature design, and attention dynamically screens for informative temporal cues; together, they stabilize model detection performance in complex, time-varying industrial scenarios.

\subsection{Self-Supervised Pre-training for Time Series}
Self-supervised pre-training mines an effective feature representation from unlabeled sequential data via auxiliary pretext tasks, mitigating the annotation-shortage bottleneck in unsupervised anomaly detection. Current relevant methods are primarily divided into the contrastive and generative learning branches.

Contrastive learning constructs paired sample constraints to regularize feature learning. CARLA \cite{Darban2025CARLA} generates positive/negative pairs via artificial anomaly injection. TimeCLR \cite{Yang2022TimeCLR} constructs matching samples based on dynamic time warping. CoST \cite{Woo2022CoST} introduces frequency-domain contrast loss to decouple trend and seasonal components. TS2Vec \cite{Yue2022TS2Vec} builds hierarchical contextual embeddings. TF-C \cite{Zhang2022TFC} enforces feature alignment across time and frequency domains. TDSRL \cite{Dai2024TDSRL} combines global dependency modeling and local contrast boundary discrimination to impose composite optimization constraints.

Masked reconstruction dominates generative self-supervised pre-training. MAET \cite{Wang2025MAET} reconstructs masked subsequences using an LSTM encoder, and TFMAE \cite{Fang2024TFMAE} builds a dual-masked autoencoder spanning time and frequency domains. Patch order restoration is another vital research direction: PPT \cite{Kim2025PPT} breaks the original temporal consistency by channel shuffling and integrates an ordering constraint with contrastive learning for model supervision. Existing sequence rearrangement-based approaches include BTSF \cite{Ling2022BTSF}, which performs dual consistency verification at the frame and segment levels, and SIM-AD \cite{Zhong2025SimAD}, which unifies contrastive learning and multi-step sequence prediction to enhance embedding discriminability.

Modern self-supervised time-series learning evolves from elementary reconstruction and pairwise contrast to multi-task joint optimization, with the modeling scope extended from the pure time domain to frequency and hybrid multidimensional spaces. Reasonably designed pretext tasks compensate for insufficient labeled data and generate high-quality embeddings to support downstream unsupervised anomaly detection.
\begin{figure*}
	\centering
	\includegraphics[width=1\textwidth]{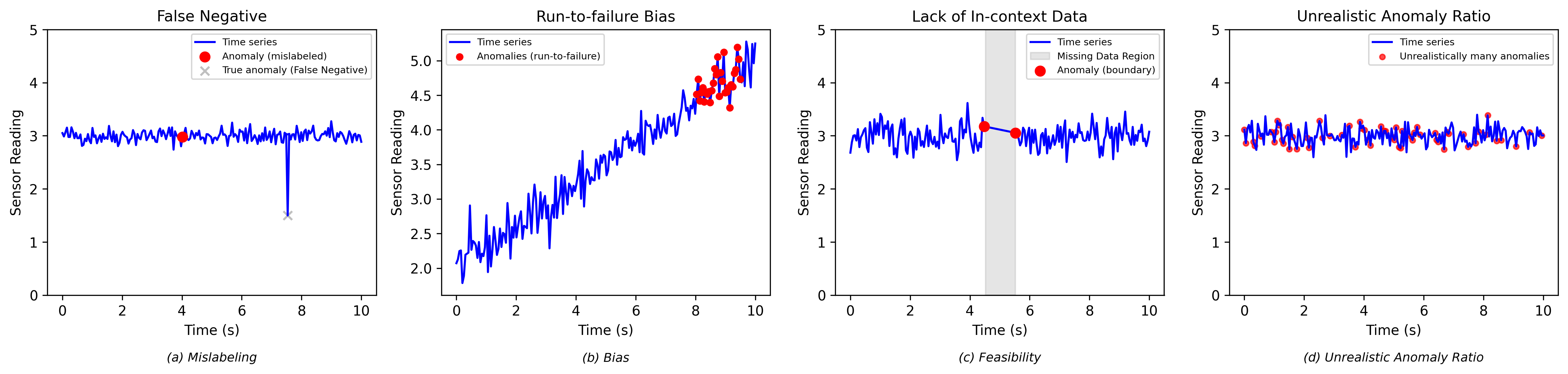}
	\caption{Classification of common defects found in the current dataset. Anomalies are marked in red.}
	\label{FIG:1}
\end{figure*}
\section{Method}
\label{sec3}
\subsection{Problem Definition}
This work addresses the task of semi-supervised time series anomaly detection. Under this paradigm, the training set consists exclusively of normal time series samples. The model learns the distribution and temporal evolution patterns of normal data to identify anomalous time points and segments with deviant behavioral patterns in test series.

Let a time series be $\boldsymbol{X} = \{\boldsymbol{x}_1, \boldsymbol{x}_2, \dots, \boldsymbol{x}_N\} \in \mathbb{R}^{d \times N}$, where $N$ denotes the total length of the series and $d$ is the dimension of observed variables, $\boldsymbol{x}_t \in \mathbb{R}^d$ represents the observation vector at time step $t$, and $t \in \{1,2,\dots,N\}$. Specifically, the time series is univariate when $d=1$, and multivariate when $d>1$.

The model processes the input time series $\boldsymbol{X}$ and outputs an anomaly score $s_t \in \mathbb{R}$ for each time step. The magnitude of the anomaly score is positively correlated with the degree of deviation from normal patterns at the corresponding time step. We define a global anomaly threshold $\theta \in \mathbb{R}$ and adopt an indicator function $I(\cdot)$ to discriminate anomalies.
\begin{equation}
	I(s_t, \theta) =
	\begin{cases}
		1, & s_t > \theta, \\
		0, & s_t \leq \theta.
	\end{cases}
\end{equation}
where $I(s_t, \theta) = 1$ indicates that the $t$-th time step is abnormal, while $I(s_t, \theta) = 0$ denotes a normal time step.

Accordingly, the sets of anomalous and normal time steps are defined as follows.
\begin{equation}
	\begin{aligned}
		\mathcal{T}_{\text{anom}} &= \{ t \mid I(s_t, \theta) = 1,\ t\in \{1,2,\dots,N\} \}, \\
		\mathcal{T}_{\text{norm}} &= \{ t \mid I(s_t, \theta) = 0,\ t\in \{1,2,\dots,N\} \}.
	\end{aligned}
\end{equation}

The corresponding anomalous and normal subsequences are formulated as
\begin{equation}
	\begin{aligned}
	\boldsymbol{X}_{\text{anom}} &= \{ \boldsymbol{x}_t \mid t \in \mathcal{T}_{\text{anom}} \}, \\
	\boldsymbol{X}_{\text{norm}} &= \{ \boldsymbol{x}_t \mid t \in \mathcal{T}_{\text{norm}} \}.
    \end{aligned}
\end{equation}

The objective of time series anomaly detection is to characterize the normal behavioral patterns embedded in time series data, compute per-step anomaly scores, and ultimately accurately identify the anomalous time set $\mathcal{T}_{\text{anom}}$ and abnormal subsequence $\boldsymbol{X}_{\text{anom}}$.

Existing mainstream benchmarks for time-series anomaly detection universally suffer from four inherent defects, as shown in Fig.~\ref{FIG:1}.

\textbf{Incorrect annotations}: Mislabeling of normal instances as anomalies and missing genuine anomalies introduce label noise and contaminate model optimization.

\textbf{Temporal position bias}: Anomalies tend to cluster at late timestamps, leading to equipment-failure-induced location bias; this bias misleads the model into fitting positional patterns rather than intrinsic anomalous characteristics.

\textbf{Insufficient contextual information}: Several abnormal samples lack adjacent normal background data, hindering the model from capturing feature discrepancies between normal and abnormal patterns.

\textbf{Unrealistic anomaly ratio}: Benchmark datasets assume an excessively high proportion of anomalous samples, which is inconsistent with the extremely sparse real-world anomaly distribution in industrial environments, leading to overestimates of empirical detection performance.

\begin{figure*}
	\centering
	\includegraphics[width=0.9\textwidth]{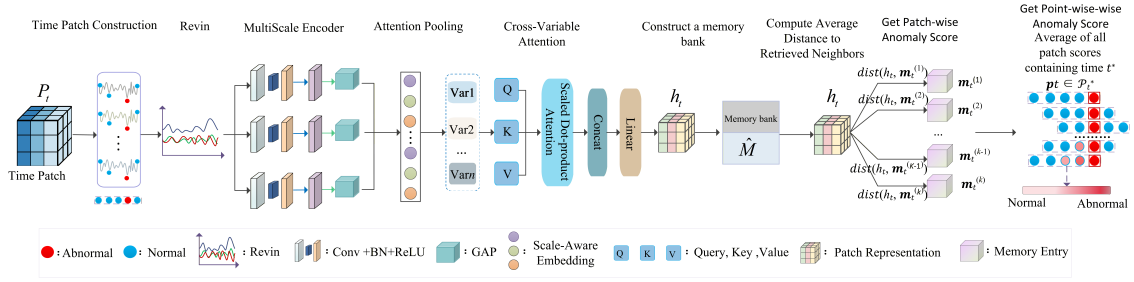}
	\caption{Architecture of the PaAno+ multivariate time-series anomaly detection system.}
	\label{FIG:2}
\end{figure*}

\subsection{Overall Framework}
The overall framework of the proposed method is illustrated in Fig.~\ref{FIG:2}. The core principle is to construct a discriminative embedding space via patch-based representation learning, in which normal temporal patterns form compact feature clusters. In contrast, anomalous patterns deviate from this regular distribution of features, thereby enabling reliable anomaly detection.

The proposed method comprises three sequential stages: patch preprocessing, encoder training, and anomaly inference.

\textbf{Patch Extraction and Preprocessing}: A sliding-window strategy is adopted to extract overlapping short-term patches $p_t = (x_t, x_{t+1}, \dots, x_{t+w-1})$ from the raw time series, where $w$ denotes the patch window size. Each extracted patch is standardized with instance normalization to improve the model's robustness to shifts in the data distribution.

\textbf{Encoder Training}: A lightweight multiscale encoder $f_\theta$ is built to project temporal patches into low-dimensional discriminative embeddings. For multivariate time series, a cross-variable attention mechanism is incorporated to explicitly model inter-variable dependencies. The encoder integrates multiscale temporal encoding with cross-variable feature fusion to enable comprehensive feature learning. The entire network is optimized using a joint loss function that combines triplet and temporal-ordering losses, with the well-trained encoder $f_\theta$.

\textbf{Anomaly Detection}: A raw memory bank $M$ is constructed from patch embeddings derived from the training set, and K-means clustering is then applied to compress $ M$ into a compact set of prototypes. In the inference stage, the K-nearest-neighbor distances between test patch embeddings and memory bank prototypes are computed for each timestamp, and the average distance is used as the final anomaly score.

\subsection{Multiscale Time-Series Encoder}
The multiscale temporal encoder is designed to capture short-range, medium-range, and long-range sequential features from time-series data. Equipped with cross-scale attention pooling and residual connections, the encoder adaptively fuses multiscale features to produce informative and comprehensive temporal representations.

The encoder adopts a three-branch parallel convolutional architecture for multiscale modeling, with kernel sizes of 3, 5, and 9 to capture features at fine, medium, and coarse scales, respectively. Each branch sequentially implements convolution, batch normalization, ReLU activation, and adaptive average pooling, yielding feature embeddings with unified dimensions.

A cross-scale attention pooling mechanism is further introduced to realize adaptive multiscale feature fusion. The output features of the three branches are concatenated along the feature dimension, and a fully connected layer is used to compute the attention weight $W_{\text{attn}, i}$ for each scale. Weighted aggregation based on the learned attention weights is finally performed to generate the refined multiscale temporal representation.
\begin{equation}
	X_{\text{fused}}=\sum_{i=1}^3 W_{\text{attn},i}\cdot X_{\text{scale},i}
	\label{eq:1}
\end{equation}
where $X_{\text{scale},i}$ denotes the output of the $i$-th scale branch. Subsequently, a residual connection is incorporated to facilitate gradient propagation and feature preservation, as formulated in Eq. \eqref{eq:2}.
\begin{equation}
	X_{\text{final}}=X_{\text{fused}}+\gamma\cdot X_{\text{residual}}
	\label{eq:2}
\end{equation}
where $\gamma$ is a learnable parameter, and $X_{\text{residual}}$ is a linear mapping of the raw outputs from each branch, as formulated in Eq. \eqref{eq:3}.
\begin{equation}
	X_{\text{residual},i}=W_{\text{residual},i}\cdot X_{\text{scale},i}
	\label{eq:3}
\end{equation}
Residual connections are embedded in each convolutional branch to retain intrinsic feature information, thereby strengthening the feature representation capability and stabilizing network optimization. The final fused feature Xfinal aggregates complementary multiscale temporal cues, yielding discriminative embeddings for downstream anomaly detection. Leveraging the multiscale architecture and adaptive attention weighting, the encoder simultaneously captures short- and long-range temporal dependencies, thereby improving its ability to model complex time-series patterns.

\subsubsection{Cross-Scale Attention Fusion}
The cross-variable fusion attention mechanism adaptively models interdependencies among multiple variables. By quantifying each variable’s contribution using multi-head self-attention, the module dynamically aggregates features weighted by attention, thereby enhancing the model’s ability to capture complex multivariate temporal correlations. Given input features $X\in\mathbb{R}^{B\times D\times T}$, where $B$, $D$, and $T$ denote the batch size, variable dimension, and time step length, respectively. The input is first projected into a latent feature space via a linear transformation.
\begin{equation}
	X_{\text{proj}} = W_{\text{proj}} X
	\label{eq:4}
\end{equation}
where $W_{\text{proj}}\in\mathbb{R}^{T\times T}$ is the transformation matrix, and $X_{\text{proj}}\in\mathbb{R}^{B\times D\times T}$ represents mapped feature. The values of $Q$, $K$, and $V$ are computed based on the mapped features, as formulated in Eq. \eqref{eq:5}.
\begin{equation}
	Q=W_Q X_{\text{proj}},\quad K=W_K X_{\text{proj}},\quad V=W_V X_{\text{proj}}
	\label{eq:5}
\end{equation}
where $W_Q$, $W_K$, and $W_V\in\mathbb{R}^{T\times T}$ are the transformation matrices for $Q$, $K$, and $V$, respectively. The attention weight matrix is obtained by applying the Softmax normalization to the dot product of $Q$ and $K$.
\begin{equation}
	A=\text{Softmax}\left(\frac{QK^\top}{\sqrt{T}}\right)
	\label{eq:6}
\end{equation}
where $A\in\mathbb{R}^{D\times D}$ represents the relative importance of the variables. The value matrix is weighted and fused using attention weights, as formulated in Eq. \eqref{eq:7}.
\begin{equation}
	X_{\text{attn}}=A V
	\label{eq:7}
\end{equation}
Concatenate the attention-weighted features with the original projected features, as formulated in Eq. \eqref{eq:8}.
\begin{equation}
	X_{\text{concat}}=\mathrm{Concat}\big(X_{\text{attn}},X_{\text{proj}}\big)
	\label{eq:8}
\end{equation}
The concatenated features contain both weighted and raw information, enhancing expressive power. Finally, the final output is obtained through a fully connected layer.
\begin{equation}
	y=W_{\text{out}} X_{\text{concat}}
	\label{eq:9}
\end{equation}
where $W_{\text{out}}\in\mathbb{R}^{T\times T}$ represents the output layer weights, and $y\in\mathbb{R}^{B\times T}$ represents the model output. This mechanism adaptively extracts interactive features among variables through multi-head self-attention, thereby enhancing the model’s ability to model multivariate time-series data.

\subsection{Pre-training Loss for Patch Order Restoration}
To enhance the encoder’s ability to model time-series dynamics, this study introduces a time-series sequence-restoration pre-task. Drawing inspiration from visual jigsaw puzzle reconstruction and adapted to the time-series context, this task requires the model to restore the original time-series order from a shuffled sequence of consecutive patches. Let the window size be $T$. For each training sample, a continuous sequence of patches $\{p_t, p_{t+1}, \dots, p_{t+T-1}\}$ is sampled from the patch set $P$. This sequence is then randomly shuffled by a permutation $\pi$ to generate a disordered sequence $\{p_{\pi(1)}, p_{\pi(2)}, \dots, p_{\pi(T)}\}$. The model learns to predict the inverse permutation $\pi^{-1}$ corresponding to the original time-series order.

This study constructs a time-series ranking head $c_\theta$ for task learning. It consists of a two-layer MLP with a hidden dimension of 512 and a dropout rate of 0.1. The concatenated embedding features of $T$ patches are fed into the ranking head, which outputs a $T \times T$ Logit matrix $\boldsymbol{O}$, where $O_{b, i,j}$ represents the confidence score of the $i$-th patch at the $j$-th time step in batch $b$. The ranking loss for pre-training is formulated as
\begin{equation}
	\mathcal{L}_{\text{ranking}} = -\frac{1}{M_{\text{valid}}}\sum_{b=1}^{M_{\text{valid}}}\sum_{i=1}^{T}\log\frac{\exp\big(O_{b,i,\pi^{-1}(i)}\big)}{\sum_{j=1}^{T}\exp\big(O_{b,i,j}\big)}
	\label{eq:10}
\end{equation}
where $M_{\text{valid}}$ denotes the number of valid samples in a batch that contain complete, consecutive $T$-patch sequences, and $\pi^{-1}$ is the inverse of the original temporal order. This pre-task loss is optimized together with the main anomaly detection loss only during the early training phase, thereby guiding the encoder to capture intrinsic temporal dependencies between adjacent patches and improving final anomaly detection performance.

\subsection{Joint Loss Function}
The overall training objective is a weighted combination of triplet loss and time-series ranking loss.
\begin{equation}
	\mathcal{L}_{\text{total}}=\mathcal{L}_{\text{triplet}}+\lambda\mathcal{L}_{\text{ranking}}
	\label{eq:11}
\end{equation}
where $\lambda$ is the weight adjustment factor for the ranking loss. The specific expressions for the two types of loss functions are given below.

Triplet loss is used to construct a highly discriminative embedding space, reducing the feature distance between similar time-series patches while increasing the feature differences between dissimilar patches. For any anchor patch $p_i$, the positive sample $p_i^+$ is generated by randomly shifting it by $\pm r$ time steps (excluding zero shifts), ensuring that it shares consistent local time-series patterns with $ p_i$. For negative samples, the $p_i^-$ strategy employs farthest-negative-sample sampling, selecting the patch in the current mini-batch $B$ with the largest cosine distance to the anchor embedding as the negative sample. Compared to random negative sample sampling, this strategy provides stronger contrast constraints, guiding the encoder to learn more discriminative feature representations.

Let the encoder $f_\theta$ map the patches to embedded features $h = f_\theta(P)$. The 256-dimensional projection head $g_\theta$ further maps the features to the contrastive learning space $z = g_\theta(h)$. The specific definition of the triplet loss is

\begin{equation}
	\mathcal{L}_{\text{triplet}}=\frac1M\sum_{i=1}^M\max\big\{\mathrm{dist}(z_i,z_i^+)-\mathrm{dist}(z_i,z_i^-)+\delta,0\big\}
	\label{eq:12}
\end{equation}
where $M$ represents the batch size, and the distance function $\mathrm{dist}(\cdot,\cdot)$ uses the cosine distance, $\mathrm{dist}(a, b) = 1 - \cos(a, b)$. $\delta = 0.5$ is the margin parameter, used to constrain the minimum difference in feature distances between positive and negative samples. This loss function requires that the feature distance between anchor points and negative samples be at least $\delta$ smaller than that between anchor points and positive samples, thereby constructing clear cluster boundaries in the embedding space. Normal time-series patches form compact feature clusters, while anomalous patches, due to their deviation from the normal time-series distribution, fall outside the cluster boundaries, thereby supporting the subsequent calculation of distance-based anomaly scores.

\begin{figure*}
	\centering
	\includegraphics[width=0.9\textwidth]{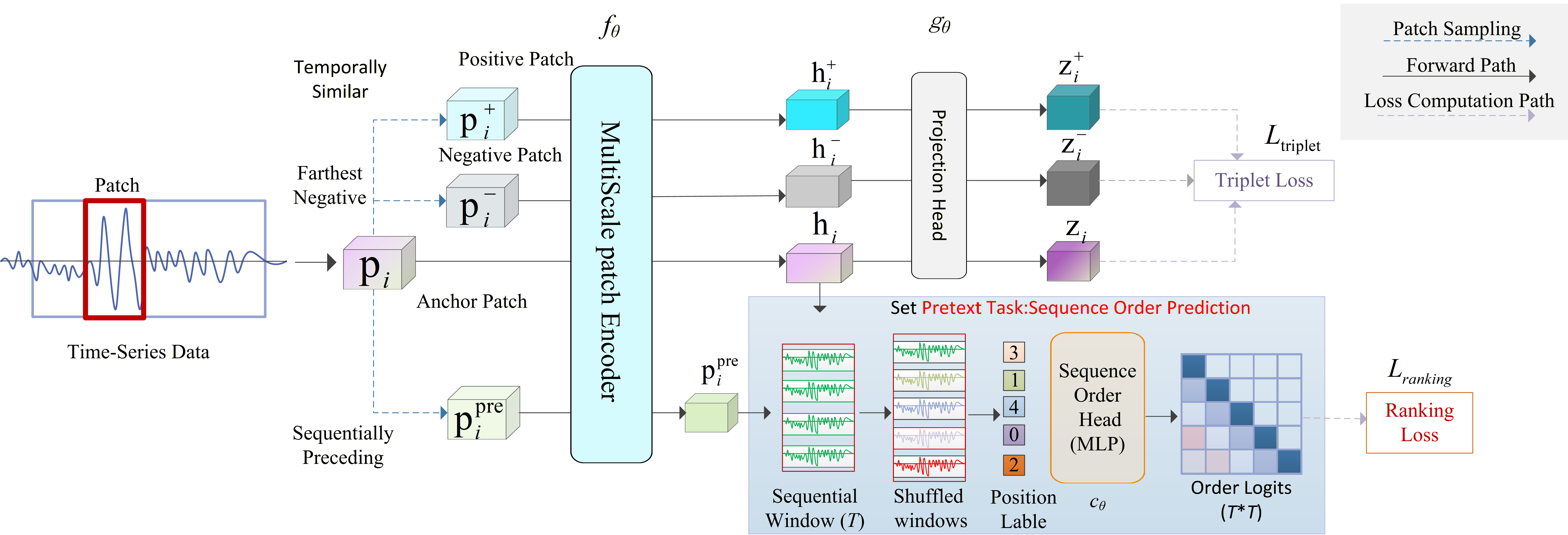}
	\caption{Training workflow of the PaAno+ model.}
	\label{FIG:3}
\end{figure*}
As illustrated in Fig. \ref{FIG:3}, the input time series is partitioned into sequential patches prior to feature extraction using the proposed multiscale encoder. Triplet loss and temporal rearrangement reconstruction loss jointly optimize the entire network. The triplet loss enforces compact feature clustering for homogeneous patches and feature dispersion across dissimilar samples. The rearrangement loss recovers the original temporal order in randomly shuffled patch sequences, thereby excavating fine-grained temporal dependencies. The two loss terms are weighted and aggregated for joint training, with the rearrangement loss weight linearly decaying during the initial training phase. After training convergence, only the learned encoder is preserved for downstream anomaly inference.

\subsection{Memory Banks and Anomaly Score Calculation}
Upon training convergence, the patch encoder $f_\theta$ constructs a discriminative embedding space in which homogeneous normal patches form compact clusters and heterogeneous normal temporal patterns are distributed across disjoint feature regions. In this work, the original memory bank $M$ is built from embedding features of all training patches.
\begin{equation}
	M=\big\{f_\theta(p_t)\,\big|\, p_t\in P\big\}
	\label{eq:13}
\end{equation}

The memory bank archives store embeddings of normal temporal patterns, which serve as the baseline for anomaly evaluation on test patches. A larger deviation between test patch embeddings and the memorized normal distribution corresponds to a higher anomaly score. To alleviate storage and computational overhead during inference, core-set sampling is adopted to compress the raw memory bank. Specifically, K-means clustering partitions the original memory bank $M$ into $K$ feature clusters, and the embedding closest to each cluster centroid is selected as the representative prototype to construct a compact, lightweight prototype memory bank.
\begin{equation}
	\hat{M}=\{m_i\}_{i=1}^K
	\label{eq:14}
\end{equation}

This compression strategy substantially reduces memory consumption while preserving the fundamental distribution coverage of normal temporal patterns. As illustrated in Fig.~\ref{FIG:4}, compressing the memory bank to 10\% of its original size results in negligible performance degradation. The model maintains stable detection accuracy even at a compression ratio of 1\%, demonstrating strong robustness to variations in memory scale. During inference, all patches $P_t$ covering each timestamp $t$ are first extracted. The patch-level anomaly score is defined as the average cosine distance between the patch embedding and its $k$ nearest prototypes retrieved from the compressed memory bank.
\begin{equation}
	S(p_t)=\frac1k\sum_{i=1}^k \mathrm{dist}\big(f_\theta(p_t),m_i^{(i)}\big)
	\label{eq:15}
\end{equation}

The final anomaly score for each timestamp is computed by averaging the scores of all patches covering that timestamp.
\begin{equation}
	S_{t_*}=\frac1{|P_{t_*}|}\sum_{p_t\in P_{t_*}} S(p_t)
	\label{eq:16}
\end{equation}

\subsection{Evaluation Criteria}
To evaluate the detection performance of PaAno+, this study selected six metrics, grouped into two main categories: interval-level and point-level. The interval-level metrics include:

VUS-PR: This metric jointly considers varying classification thresholds and temporal shift tolerance, which comprehensively evaluate the model’s overall anomaly recognition accuracy and abnormal interval localization capability. Its formal definition is given as
\begin{equation}
	\text{VUS-PR}=\iint_{\tau\in[0,1],\delta\in[0,\Delta]} \text{PR}(\tau,\delta)\mathrm{d}\tau\mathrm{d}\delta
	\label{eq:17}
\end{equation}
where $\tau$ represents the anomaly detection threshold, $\delta$ is the time-series tolerance offset, $\Delta$ is the upper limit of the offset, and $\text{PR}(\tau, \delta)$ is the precision-recall curve for the corresponding parameters.

VUS-ROC: This metric evaluates model ranking performance across diverse decision thresholds under temporal tolerance shifts, with its formal definition presented as
\begin{equation}
	\text{VUS-ROC}=\iint_{\tau\in[0,1],\delta\in[0,\Delta]} \text{ROC}(\tau,\delta)\mathrm{d}\tau\mathrm{d}\delta
	\label{eq:18}
\end{equation}

Range-F1: Precision and recall are quantified by the interval overlap between predictions and the ground truth, with the peak F1 value used as the final score. Let $\hat{S}$ and $S$ denote the sets of predicted and ground-truth anomalous intervals, respectively. Its formal definition is expressed as
\begin{equation}
	\text{Range-F1}=\max_{\theta}\frac{2\cdot \text{Precision}_\theta \cdot \text{Recall}_\theta}{\text{Precision}_\theta+\text{Recall}_\theta}
	\label{eq:19}
\end{equation}
where $\theta$ denotes the anomaly decision threshold, and precision and recall are computed using the overlap in duration between predicted and ground-truth anomalous intervals. Point-level metrics include:

AUC-PR: This metric measures the area under the precision-recall (PR) curve and is robust against class imbalance inherent to time-series data, whose formal definition is given as
\begin{equation}
	\text{AUC-PR}= \int_{0}^{1} P(R)\mathrm{d}R
	\label{eq:16}
\end{equation}
where $R$ represents the recall rate, and $P(R)$ is the precision corresponding to that recall rate.

AUC-ROC: This metric calculates the area under the receiver operating characteristic (ROC) curve, which quantifies the model's overall ranking performance.
\begin{equation}
	\text{AUC-ROC}= \int_{0}^{1}\text{TPR}(\text{FPR})\mathrm{d}(\text{FPR})
	\label{eq:20}
\end{equation}

Point-F1: This metric computes the optimal F1 score based on point-level anomaly detection results across varying classification thresholds.
\begin{equation}
	\text{Point-F1}= \max_{\tau}\frac{2\cdot P_\tau \cdot R_\tau}{P_\tau+R_\tau}
	\label{eq:21}
\end{equation}
where $\tau$ denotes the threshold for pointwise classification, and $P_\tau$ and $R_\tau$ correspond to the resulting precision and recall at that threshold, respectively.

Among the above metrics, the first four are threshold-independent. Notably, all six metrics eliminate pointwise correction operations, thereby preventing evaluation bias. In this work, VUS-PR is adopted as the primary evaluation indicator, while the other five metrics serve as supplementary references. Higher values across all metrics indicate superior anomaly detection performance.

\section{Experimentd}
\label{sec4}
This section presents a comprehensive empirical validation of the proposed method for time-series anomaly detection. It first specifies the experimental datasets, evaluation metrics, and comparative baseline methods, then elaborates on the implementation configurations of the proposed model, and finally provides quantitative results for univariate and multivariate detection tasks.

\subsection{Dataset}
This study uses the TSB-AD benchmark \cite{Liu2024Benchmark} as the experimental dataset, which is designed to address the inherent limitations of conventional evaluation protocols for time-series anomaly detection. The overall statistical attributes of the benchmark are summarized in Table~\ref{tab:1}. This benchmark consists of two subsets: the univariate TSB-AD-U and the multivariate TSB-AD-M. Specifically, TSB-AD-U includes 350 test sequences with an average length of 51,886.7 timesteps and an anomaly ratio of 4.5\%. In comparison, TSB-AD-M comprises 180 test sequences with an average length of 108,826.7 timesteps and an anomaly ratio of 5.0\%. All sequences are pre-partitioned into training and test splits, where the training split only contains normal samples and the test split is annotated with accurate anomaly labels.

The TSB-AD benchmark eliminates prevalent flaws in raw datasets, including inconsistent annotations and distorted anomaly distributions, and excludes point-adjustment operations that may introduce evaluation bias. Accordingly, it provides a fair and unified evaluation platform for comprehensive comparison among different anomaly detection methods. A data illustration of the benchmark is shown in Table~\ref{tab:1}. Each subset is further divided into a tuning set and an evaluation set, used for hyperparameter optimization and final performance validation, respectively. All time-series samples have fixed segmentation boundaries, and data segments before the boundaries are designated as training samples.

\begin{table}[htbp]
	\centering
	\caption{Statistics of the TSB-AD benchmark.}
	\label{tab:1}
	\scriptsize
	\setlength{\tabcolsep}{6pt}
	\begin{tabular}{lllllll}
		\toprule
		Category & Split & \#TS & \begin{tabular}{@{}c@{}}Avg Length\end{tabular} & \begin{tabular}{@{}c@{}}Avg Anomaly Length\end{tabular} & \begin{tabular}{@{}c@{}}Avg \#Anomalies\end{tabular} & \begin{tabular}{@{}c@{}}Anomaly Ratio\end{tabular}\\
		\midrule
		\multirow{3}{*}{TSB-AD-U} & All    & 870 & 38814.1  & 179.5 & 39.7  & 2.4\% \\
		& Eval   & 350 & 51886.7  & 321.3 & 46.6  & 4.5\% \\
		& Tuning & 48  & 47143.3  & 185.9 & 82.6  & 3.5\% \\
		\midrule 
		\multirow{3}{*}{TSB-AD-M} & All    & 200 & 107760.4 & 582.6 & 71.1  & 5.1\% \\
		& Eval   & 180 & 108826.7 & 591.2 & 67.7  & 5.0\% \\
		& Tuning & 20  & 98164.1  & 504.7 & 101.1 & 5.7\% \\
		\bottomrule
	\end{tabular}
\end{table}

\subsection{Complexity Analysis}
This section evaluates the computational overhead of PaAno+ under univariate and multivariate time series scenarios in terms of time complexity, space complexity, and model parameter count, as shown in Table \ref{tab:complexity_compare}. The sliding step size is fixed at $\text{step}=96$ across all experiments; the time series window length is set to $T=2$ for the univariate setup and $T=5$ for the multivariate setup. Table \ref{tab:complexity_compare} summarizes the complexity metrics for the two scenarios. The two variants share identical asymptotic orders of time and space complexity. The multivariate variant incurs only a marginal increase in the number of parameters due to the longer time window and cross-variable interaction modules, and the model remains lightweight overall. Specifically, $N_\textit{MS}$ denotes the parameter count of the multi-scale encoder, $N_\textit{CV}$ that of the cross-variable attention module, $N_\textit{PH}$ that of the projection head, and $N_\textit{OH}$ that of the temporal ordering head.

\begin{table}[htbp]
	\centering
	\caption{Complexity Comparison.}
	\label{tab:complexity_compare}  
	\setlength{\tabcolsep}{6pt}
	\begin{tabular}{lll}
		\hline
		& Univariate $(T=2)$ & Multivariate $(T=5)$ \\
		\hline
		Time   & $O(ntin)$          & $O(ntin)$            \\
		Space  & $O(ntin)$          & $O(ntin)$            \\
		Param  & $\begin{aligned}
			N_U &= N_{MS}+N_{CV}+N_{PH}+N_{OH}(2)
		\end{aligned}$
		& $\begin{aligned}
			N_M &= N_{MS}+N_{CV}+N_{PH}+N_{OH}(5)
		\end{aligned}$ \\
		\hline
	\end{tabular}
\end{table}

\subsection{Baseline Method}
To comprehensively validate the detection performance of the proposed PaAno+, this study includes all baseline methods reported in the TSB-AD benchmark's univariate and multivariate evaluation protocols, as listed in Tables~\ref{tab:2} and~\ref{tab:3}. These baselines span three mainstream technical paradigms: statistical machine learning, conventional deep neural networks, and Transformer-based architectures.

In univariate anomaly detection tasks, the baseline methods employed include statistical and machine learning approaches such as SAND \cite{Boniol2021SAND}, DLinear \cite{Zeng2023TransEff} and NLinear \cite{Zeng2023TransEff}; neural network methods such as TimesNet \cite{Wu2023TimesNet}, TranAD \cite{Tuli2022TranAD}, FITS \cite{Xu2024FITS}, DADA \cite{Shentu2025DADA}, KAN-AD \cite{Paparrizos2015kshape}, PaAno \cite{Park2026PaAno}, CrossAD \cite{Li2025CrossAD} and other neural network methods; as well as AnomalyTransformer \cite{Xu2022AnomalyTrans}, DCdetector \cite{Yang2023DCDetector}, LagLlama \cite{Rasul2023LagLlama}, OFA \cite{Su2019SRNNAD}, PatchTST \cite{Nie2023Word64}, iTransformer \cite{Liu2024iTransformer}, MOMENT (FT/ZS) \cite{Goswami2024MOMENT}, TimesFM \cite{Das2024DecoderTS}, Chronos \cite{Ansari2024Chronos} and other Transformer and foundational model-based methods.

In multivariate anomaly detection tasks, baseline methods include: statistical and machine learning methods such as DLinear \cite{Zeng2023TransEff} and NLinear \cite{Zeng2023TransEff}; neural network methods such as TimesNet \cite{Wu2023TimesNet}, TranAD \cite{Tuli2022TranAD}, FITS \cite{Xu2024FITS}, DADA \cite{Shentu2025DADA}, KAN-AD \cite{Paparrizos2015kshape}, CrossAD \cite{Li2025CrossAD}, CATCH \cite{Wu2025CATCH}; and Transformer and foundational model-based methods such as AnomalyTransformer \cite{Xu2022AnomalyTrans}, DCdetector \cite{Yang2023DCDetector}, PatchTST \cite{Nie2023Word64}, OFA \cite{Su2019SRNNAD}, and iTransformer \cite{Liu2024iTransformer}.

The selected baseline suite includes traditional statistical methods and state-of-the-art deep learning models, enabling a thorough and fair evaluation of PaAno+'s effectiveness and superiority across diverse technical paradigms.

\subsection{Implementation Details}
The multiscale encoder is composed of three parallel convolutional branches with kernel sizes 3, 7, and 15. The first layer is configured with a hidden dimension of 64, and the second layer yields 128-dimensional features followed by adaptive average pooling. Branch outputs are weighted and aggregated, then concatenated with raw branch features, with the first 256 dimensions reserved. A residual linear projection unifies feature dimensions, and the encoder finally outputs a 256-d patch embedding.

The cross-variable fusion attention module is exclusively activated for multivariate inputs. It reshapes the 256-d input into a product of the variable count and the per-variable feature dimension, then applies two-headed self-attention with a dropout rate of 0.1. Attention outputs are concatenated with original variable features, and a linear layer projects the concatenated result back to 256 dimensions.

The projection head is implemented as a two-layer MLP with 256-d output. The temporal ordering head receives input of size projection dimension $\times$ window length and produces an $L \times L$ logit matrix ($L$ denotes the window length) to supervise the patch rearrangement pretext task.

Patch configuration for the ordering pretext task varies across univariate and multivariate settings: univariate tasks use a patch length of 64 with two successive windows, while multivariate tasks use a patch length of 96 and five successive windows. The ordering-loss weighting coefficient $\lambda$ linearly decays from 1 to 0 over the initial 20 training epochs and is discarded thereafter. The learning rate follows cosine annealing, ranging from $1\times10^{-3}$ to $1\times10^{-4}$, and the total number of training epochs is set to 200.

All implementations are developed in PyTorch using the Adam optimizer and a batch size of 512. The triplet margin is fixed at $\delta = 0.5$, and the offset radius for positive sample construction is set to 2. Upon convergence, K-means is run on embeddings of all normal training patches, and the top 10\% of cluster centroids are retained to build a compact memory prototype set. During inference, each test patch is matched to three memory prototypes via cosine similarity, and the average distance is used as the patch-wise anomaly score. The final pointwise anomaly score is computed by averaging the scores across all patches that cover the corresponding timestamp. All empirical experiments are conducted on a server mounted with an NVIDIA RTX 4090 GPU.

\subsection{ Patch Size Sensitivity Analysis}
Table~\ref{tab:2} reports the patch-size sensitivity experiments carried out on the TSB-AD tuning subset. Overall detection accuracy varies only slightly with patch length, confirming that on TSB-AD-U and TSB-AD-M, PaAno delivers feasible performance at a patch size of 32, with marginal accuracy gains as patch length increases. For univariate detection, patch sizes of 64 and 96 yield comparable results and outperform sizes of 32 and 128. In the multivariate setting, 64 and 96 are optimal candidates; patch length 96 marginally exceeds 64 on AUC-ROC and Point-F1, whereas excessively small or large patch sizes lead to significant performance deterioration. Based on the above tuning results, the patch size is set to 96 for both univariate and multivariate experiments in subsequent evaluations.

\begin{table}[htbp]
	\centering
	\caption{Sensitivity analysis of patch size. The optimal values for each metric in the table are bolded, and the next-best values are underlined. All experimental results are expressed as percentages.}
	\label{tab:2}
	\scriptsize
	\setlength{\tabcolsep}{8pt} 
	\begin{tabular}{lccccccc}
		\toprule
		Model & Size & VUS-PR & VUS-ROC & Range-F1 & AUC-PR & AUC-ROC & Point-F1 \\
		\midrule
		\multirow{4}{*}{TSB-AD-U}
		&32 &42.7 &86.2 &45.2 &36.5 &82.2 &42.2\\
		&64 &\underline{46.4} &89.1 &\underline{46.5} &\textbf{42.4} &86.3 &\underline{47.4}\\
		&96 &\textbf{46.4} &\underline{90.2} &\textbf{48.9} &\underline{41.4} &\underline{87.7} &\textbf{48.0}\\
		&128&46.1 &\textbf{90.7} &45.6 &38.7 &\textbf{88.4} &45.5\\
		\midrule
		\multirow{4}{*}{TSB-AD-M}
		&32 &38.8 &80.0 &32.9 &29.5 &71.6 &34.5\\
		&64 &\underline{47.6} &\underline{85.0} &\underline{39.1} &37.4 &\underline{78.8} &43.0\\
		&96 &\textbf{47.6} &\textbf{85.0} &38.9 &\underline{37.5} &\textbf{78.9} &\textbf{43.2}\\
		&128&47.3 &82.7 &\textbf{39.8} &\textbf{38.5} &77.2 &\underline{43.1}\\
		\bottomrule
	\end{tabular}
\end{table}

\subsection{Quantitative Results}
\subsubsection{Univariate Time Series Anomaly Detection}
Table~\ref{tab:3} summarizes quantitative results of all competing approaches evaluated on the TSB-AD-U test set. The proposed PaAno+ achieves optimal scores across all six evaluation metrics (i.e., VUS-PR = 0.55, VUS-ROC = 0.90, Range-F1 = 0.52, AUC-PR = 0.49, AUC-ROC = 0.87, and Point-F1 = 0.54) and comprehensively outperforms the original PaAno and all comparative baselines. Among the statistical machine learning baselines, SNAD performed relatively well, but its overall metrics still fell short of the proposed method. Among the deep network baselines, CrossAD ranked second in overall performance, with a VUS-PR of only 0.45. This model has 0.9 million parameters, which is lower than the 1.1 million in PaAno+, and there is a significant gap in detection accuracy. Various Transformer-based models and foundational large language models performed relatively poorly overall; AnomalyTransformer and DCDetector achieved VUS-PRs of only 0.12 and 0.09, respectively, whilst their parameter counts were significantly higher than that of PaAno+. The proposed PaAno+ achieves detection performance superior to the vast majority of Transformer-based approaches with a parameter count of just 1.1 million, demonstrating that the model successfully balances lightness with detection accuracy.

\begin{table*}[htbp]
	\centering
	\caption{Performance of each method on the TSB-AD-U Eval dataset.}
	\label{tab:3}
	\small
	\setlength{\tabcolsep}{1pt}
	\begin{tabular}{llcccccccc}
		\toprule
		\multicolumn{2}{c}{Method} & \multicolumn{3}{c}{Range-Wise Measure$\uparrow$} & \multicolumn{3}{c}{Point-Wise Measure$\uparrow$} & \multicolumn{2}{c}{Computational Cost $\downarrow$} \\
		\cmidrule(lr){3-5}\cmidrule(lr){6-8}\cmidrule(lr){9-10}
		& & VUS-PR & VUS-ROC & Range-F1 & AUC-PR & AUC-ROC & Point-F1 & \#Params &Time \\
		\midrule
		\multirow{3}{*}{Stat \& ML}
		& SAND \cite{Boniol2021SAND} (2022)&0.34&0.76&0.36&0.29&0.73&0.35&0.2M&--\\
		& DLinear \cite{Zeng2023TransEff} (2023)&0.25&0.74&0.22&0.21&0.62&0.26&$<$0.1M&2.9s\\
		& NLinear \cite{Zeng2023TransEff} (2023)&0.23&0.72&0.20&0.18&0.62&0.23&$<$0.1M&5.8s\\
		\midrule
		\multirow{6}{*}{NN}
		& TimesNet \cite{Wu2023TimesNet} (2022)&0.26&0.72&0.21&0.18&0.61&0.24&$<$0.1M&11.2s\\
		& FITS \cite{Xu2024FITS} (2023)&0.26&0.73&0.20&0.17&0.61&0.23&$<$0.1M&3.1s\\
		& DADA \cite{Shentu2025DADA} (2025)&0.31&0.77&0.31&0.29&0.71&0.38&1.84M&0.8s\\
		& KAN-AD \cite{Paparrizos2015kshape} (2025)&0.43&0.82&0.43&0.41&0.80&0.44&$<$0.1M&12.1s\\
		& CrossAD \cite{Li2025CrossAD} (2025)&0.45&0.84&0.41&0.44&0.78&0.47&0.9M&175.8s\\
		\midrule
		\multirow{9}{*}{Transformer}
		& TranAD \cite{Tuli2022TranAD} (2022)&0.26&0.68&0.25&0.20&0.57&0.25&1.1M&56.2s\\
		& AT \cite{Xu2022AnomalyTrans} (2022)&0.12&0.56&0.14&0.08&0.50&0.12&4.8M&48.9s\\
		& DCdetector \cite{Yang2023DCDetector} (2023)&0.09&0.56&0.10&0.05&0.50&0.10&0.9M&5.8s\\
		& Lag-Llama \cite{Rasul2023LagLlama} (2023)&0.27&0.72&0.31&0.25&0.65&0.30&2.5M&1220.8s\\
		& OFA \cite{Su2019SRNNAD} (2023)&0.24&0.71&0.20&0.16&0.59&0.22&0.6M&171.1s\\
		& iTransformer \cite{Liu2024iTransformer} (2024)&0.22&0.74&0.18&0.16&0.61&0.21&0.5M&9.8s\\
		& MOMENT (FT) \cite{Goswami2024MOMENT} (2024)&0.39&0.76&0.35&0.30&0.69&0.35&109.6M&43.6s\\
		& MOMENT (ZS) \cite{Goswami2024MOMENT} (2024)&0.38&0.75&0.36&0.30&0.68&0.35&109.6M&42.9s\\
		& TimesFM \cite{Das2024DecoderTS} (2024)&0.30&0.74&0.34&0.28&0.67&0.34&203.5M&83.8s\\
		& Chronos \cite{Ansari2024Chronos} (2024)&0.27&0.73&0.33&0.26&0.66&0.32&8.0M&56.8s\\
		\midrule
		& PaAno \cite{Park2026PaAno} (2026)&0.51&0.88&0.48&0.46&0.85&0.51&0.3M&6.9s\\
		& PaAno+ (Ours) &\textbf{0.55}&\textbf{0.90}&\textbf{0.52}&\textbf{0.49}&\textbf{0.87}&\textbf{0.54}&1.1M&12.9s\\
		\bottomrule
	\end{tabular}
\end{table*}

\begin{table*}[htbp]
	\centering
	\caption{Performance of each method on the TSB-AD-M Eval dataset.}
	\label{tab:4}
	\small
	\setlength{\tabcolsep}{1pt}
	\begin{tabular}{llcccccccc}
		\toprule
		\multicolumn{2}{c}{Method} & \multicolumn{3}{c}{Range-Wise Measure$\uparrow$} & \multicolumn{3}{c}{Point-Wise Measure$\uparrow$} & \multicolumn{2}{c}{Computational Cost $\downarrow$} \\
		\cmidrule(lr){3-5}\cmidrule(lr){6-8}\cmidrule(lr){9-10}
		& & VUS-PR & VUS-ROC & Range-F1 & AUC-PR & AUC-ROC & Point-F1 & \#Params & Time \\
		\midrule
		\multirow{2}{*}{Stat \& ML}
		& DLinear  \cite{Zeng2023TransEff} (2023)&0.29&0.70&0.26&0.27&0.66&0.32&$<$0.1M&14.8s\\
		& NLinear  \cite{Zeng2023TransEff} (2023)&0.29&0.70&0.28&0.24&0.65&0.31&$<$0.1M&15.0s\\
		\midrule
		\multirow{5}{*}{NN}
		& TimesNet \cite{Wu2023TimesNet} (2022)&0.19&0.64&0.17&0.13&0.56&0.20&$<$0.1M&52.1s\\
		& FITS \cite{Xu2024FITS} (2023)&0.21&0.66&0.16&0.15&0.58&0.22&$<$0.1M&16.7s\\
		& DADA \cite{Shentu2025DADA} (2025)&0.31&0.73&0.25&0.31&0.69&0.35&1.84M&2.1s\\
		& KAN-AD \cite{Paparrizos2015kshape} (2025)&0.41&0.75&\textbf{0.41}&\textbf{0.38}&0.73&0.42&$<$0.1M&31.9s\\
		& CrossAD \cite{Li2025CrossAD} (2025)&0.33&0.77&0.37&0.34&0.74&0.38&0.9M&194.7s\\
		\midrule
		\multirow{7}{*}{Transformer}
		& TranAD \cite{Tuli2022TranAD} (2022)&0.18&0.65&0.21&0.14&0.59&0.21&1.1M&50.4s\\
		& AT \cite{Xu2022AnomalyTrans} (2022)&0.12&0.57&0.14&0.07&0.52&0.12&4.8M&55.8s\\
		& DCdetector \cite{Yang2023DCDetector} (2023)&0.10&0.56&0.10&0.06&0.50&0.10&0.9M&15.0s\\
		& OFA \cite{Su2019SRNNAD} (2023)&0.21&0.63&0.17&0.15&0.55&0.21&81.9M&532.9s\\
		& PatchTST \cite{Nie2023Word64} (2023)&0.28&0.71&0.26&0.26&0.65&0.32&0.5M&66.9s\\
		& iTransformer \cite{Liu2024iTransformer} (2024)&0.29&0.70&0.23&0.23&0.63&0.28&0.6M&24.4s\\
		& CATCH \cite{Wu2025CATCH} (2025)&0.30&0.73&0.27&0.24&0.67&0.30&210.8M&40.1s\\
		\midrule
		& PaAno \cite{Park2026PaAno} (2026)&\textbf{0.43}&0.79&\textbf{0.41}&\textbf{0.38}&0.76&\textbf{0.43}&0.3M&12.8s\\
		& PaAno+ (Ours) &0.42&\textbf{0.80}&\textbf{0.41}&\textbf{0.38}&\textbf{0.77}&\textbf{0.43}&1.5M&23.5s\\
		\bottomrule
	\end{tabular}
\end{table*}

\subsubsection{ Multivariate Time Series Anomaly Detection}
Table~\ref{tab:4} presents the quantitative results for all competing methods on the TSB-AD-M test set. PaAno+ retains state-of-the-art performance for multivariate anomaly detection: its VUS-ROC and AUC-ROC attain 0.80 and 0.77, moderately exceeding the corresponding 0.79 and 0.76 of vanilla PaAno. At the same time, the remaining evaluation metrics are comparable to PaAno. Both approaches surpass alternative baseline methods by a noticeable margin. With a total parameter budget of 1.5M, PaAno+ introduces marginal parameter overhead via the embedded multiscale encoder and cross-variable attention module. Yet, these two components substantially strengthen the discriminative capability toward multivariate feature patterns.

Among the deep network baselines, KAN-AD achieved a VUS-PR of 0.41, ranking third behind PaAno and PaAno+. Among Transformer-based methods, CATCH performed best (VUS-PR = 0.30), but its parameter count was as high as 210.8 million, far exceeding the 1.5 million of PaAno+; AnomalyTransformer and DCdetector achieved VUS-PRs of only 0.12 and 0.10, respectively, with detection performance significantly inferior to that of our method. Among statistical and machine learning methods, both DLinear and NLinear achieved a VUS-PR of 0.29, showing a clear gap compared to PaAno+.

Overall empirical evidence verifies that PaAno+ matches or moderately exceeds vanilla PaAno on multivariate detection benchmarks. Benefiting from far fewer learnable parameters than prevalent large Transformer architectures, the proposed framework achieves competitive detection precision, validating the effectiveness and lightweight merits of the multiscale encoder, cross-variable attention, and the temporal rearrangement pretext task under multivariate modeling settings.

\subsection{Ablation Experiment}
To verify the individual efficacy of three core components in PaAno+, namely the multiscale temporal encoder, cross-variable fusion attention, and temporal rearrangement pretext task, ablation studies are implemented on the evaluation splits of TSB-AD-U and TSB-AD-M. All ablated variants inherit the original hyperparameter configuration and experimental pipeline, with VUS-PR as the primary evaluation metric.

\begin{table*}[htbp]
	\centering
	\caption{Ablation experiment results. The optimal results for each metric are indicated in bold. All experimental values are presented as percentages (\%).}
	\label{tab:5}
	\small
	\setlength{\tabcolsep}{8pt}
	\begin{tabular}{llcccccc}
		\toprule
		Dataset & Ablation Variant & VUS-PR & VUS-ROC & Range-F1 & AUC-PR & AUC-ROC & Point-F1 \\
		\midrule
		\multirow{4}{*}{TSB-AD-U}
		& w/o MultiScale &54.2&\textbf{89.9}&49.6&47.8&\textbf{87.7}&52.9\\
		& w/o dual-scale &54.0&89.6&51.9&48.1&87.4&52.7\\
		& w/o pretext\_loss &52.8&88.7&50.5&46.4&86.5&51.8\\
		& PaAno+ (Ours) &\textbf{54.6}&89.6&\textbf{51.9}&\textbf{48.7}&87.5&\textbf{53.7}\\
		\midrule
		\multirow{5}{*}{TSB-AD-M}
		& w/o MultiScale &39.8&78.4&38.3&35.3&75.3&40.8\\
		& w/o dual-scale &41.0&79.4&40.4&36.2&66.3&42.3\\
		& w/o Attention &41.4&79.7&40.3&37.4&76.6&42.3\\
		& w/o pretext\_loss &39.1&78.0&38.4&34.3&74.7&40.2\\
		& PaAno+ (Ours) &\textbf{42.3}&\textbf{79.8}&\textbf{41.0}&\textbf{37.5}&\textbf{76.7}&\textbf{43.1}\\
		\bottomrule
	\end{tabular}
\end{table*}

\subsubsection{Validity of Multiscale Encoders}
To validate the efficacy of the parallel-branch configuration and the cross-scale fusion mechanism within the multiscale encoder, two model variants are constructed for an ablation study. The first variant eliminates the multiscale structure and adopts four stacked serial convolutional layers (w/o Multiscale). The second variant retains dual parallel convolutional branches and cross-scale attention fusion (w/o dual-scale).

As quantified in Table~\ref{tab:5}, on the univariate TSB-AD-U dataset, the full three-scale PaAno+ model achieves a VUS-PR of 54.6\%, while the w/o Multiscale variant declines to 54.2\% and the two-scale variant drops to 54.0\%. The two-scale structure yields a 0.2\% performance improvement over the non-multiscale structure, and the three-scale structure further improves the metric by 0.6\% compared with the two-scale counterpart. On the multivariate TSB-AD-M dataset, the full model achieves a VUS-PR of 42.3\%, whereas the w/o Multiscale variant suffers a 2.5\% drop to 39.8\%, and the two-scale variant achieves 41.0\% with a 1.3\% drop. The three-scale structure outperforms the two-scale structure by 1.3\% on multivariate tasks.

The experimental observations lead to three key conclusions. First, the multiscale encoder delivers stronger performance gains on multivariate tasks, as anomalies in multivariate time series exhibit richer multiscale temporal structure. Second, the three-scale architecture exhibits distinct superiority over the two-scale counterpart. Parallel branches with diverse receptive fields can comprehensively capture local and global anomalous temporal patterns, compensating for the two-scale structure's insufficient feature coverage. Third, the cross-scale attention fusion mechanism facilitates the interaction and complementary integration of multiscale features, thereby effectively strengthening the model's feature representation capability. Overall, the three-branch parallel convolution architecture, integrated with cross-scale attention fusion, has proven to be both rational and effective.

\subsubsection{Cross-Variable Attention Contributions}
To explore the contribution of cross-variable fusion attention to multivariate anomaly detection, an ablated variant (w/o Attention) is established by removing the cross-variable attention module while preserving the multiscale encoder and temporal ordering pretext task. The corresponding experimental results are presented in Table~\ref{tab:5}. On the TSB-AD-M dataset, removing cross-variable attention reduces the model’s VUS-PR from 42.3\% to 41.4\% (a 0.9\% drop). It decreases Point-F1 from 43.1\% to 42.3\% (a 0.8\% drop), with consistent declines across other evaluation metrics. These results confirm that cross-variable attention enhances the model’s ability to capture coupling anomalies across variables by explicitly modeling inter-variable dependencies. Since univariate time series contain no additional variable dimensions, this module is disabled for univariate tasks. Therefore, cross-variable attention is a core component enabling PaAno+ to achieve superior performance in multivariate time-series anomaly detection.

\subsubsection{The Role of Time-series Sorting Tasks}
To verify the effectiveness of the proposed temporal sequence reconstruction pretext task, a comparative variant (w/o pretext\_loss) is designed that replaces the proposed window-level sequence reconstruction task with the patch-adjacency binary classification task used by vanilla PaAno. This variant only identifies the temporal adjacency between paired patches, without reconstructing the overall temporal order of the shuffled windows. As shown in Table~\ref{tab:5}, the full PaAno+ model achieves a VUS-PR of 54.6\%, while the variant with a modified supervision task drops to 52.8\%, resulting in a 1.8\% performance degradation. For the TSB-AD-M dataset, the full model's VUS-PR of 42.3\% decreases to 39.1\% after task replacement, a 3.2\% degradation.

These results demonstrate that the temporal ordering reconstruction task provides more rigorous and structured supervision signals than patch-level binary classification. The window ordering task forces the model to recover the inherent temporal order of shuffled windows, driving the network to learn global temporal dependencies. In contrast, patch adjacency binary classification focuses only on local temporal relationships and fails to capture long-range temporal correlations. The performance degradation is more pronounced in multivariate scenarios, attributed to the dual coupling of temporal dependencies and inter-variable correlations in multivariate time series. Such complex characteristics rely heavily on comprehensive ordering supervision signals to optimize discriminative feature learning.

\subsection{ Hyperparameter Sensitivity Analysis}
This section conducts a hyperparameter sensitivity analysis across four core configurations of PaAno+: memory compression ratio, multiscale convolutional kernel combination, number of nearest neighbors $k$, and temporal window length $T$. All experiments are performed on the evaluation splits of the TSB-AD-U and TSB-AD-M datasets.
\begin{figure*}
	\centering
	\includegraphics[width=1.0\textwidth]{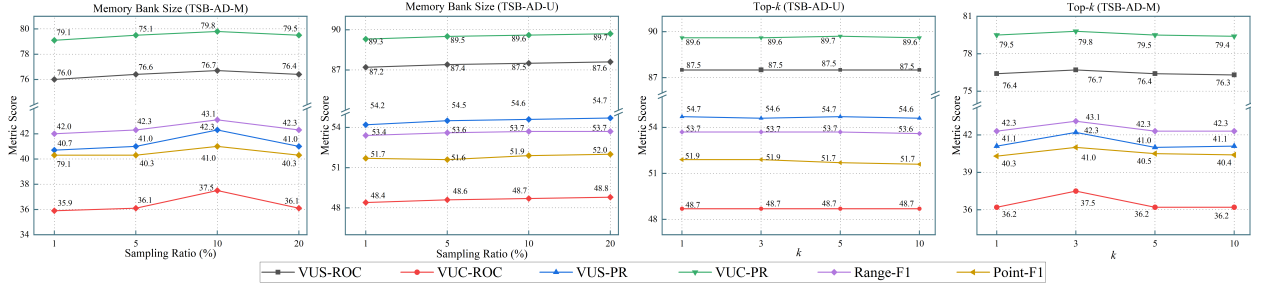}
	\caption{ Parameter sensitivity analysis of the PaAno+ model’s Top-$k$ values and memory bank size on the TSB-AD-U and TSB-AD-M Eval datasets.}
	\label{FIG:4}
\end{figure*}
PaAno+ exhibits superior robustness to hyperparameters and maintains stable detection performance across diverse parameter settings. Fig.~\ref{FIG:4} shows the sensitivity results for the memory bank compression ratio and the number of nearest neighbors. In practical industrial scenarios, time-series data features often exhibit slow distributional drift due to dynamic environmental changes. To address this issue, PaAno+ integrates a training-free online memory bank updating mechanism based on queue management. This strategy continuously updates normal feature patches and removes outdated historical features, ensuring the memory bank stays in sync with the real-time data distribution.

The nearest-neighbor number $k$ and the memory compression ratio are two critical control parameters for the updating mechanism. Model performance fluctuates slightly when $k$ varies from 1 to 5. The setting $k = 3$ achieves a favorable trade-off between detection accuracy and operational stability and is adopted as the default configuration for online inference. When the memory compression ratio decreases from 10\% to 1\%, the overall performance degradation is less than 1.5\%, enabling flexible configuration in line with practical hardware storage constraints. The maximum memory bank capacity is determined by the compression ratio, formulated as $r\times N$, where $N$ denotes the total number of original patches. During feature updates, the model maintains a fixed memory scale via a first-in, first-out queue management scheme combined with periodic resampling and K-means reclustering. This design empowers the model to adapt to non-stationary temporal data without repeated retraining, effectively reducing redundant computational overhead.

\begin{figure*}
	\centering
	\includegraphics[width=1.0\textwidth]{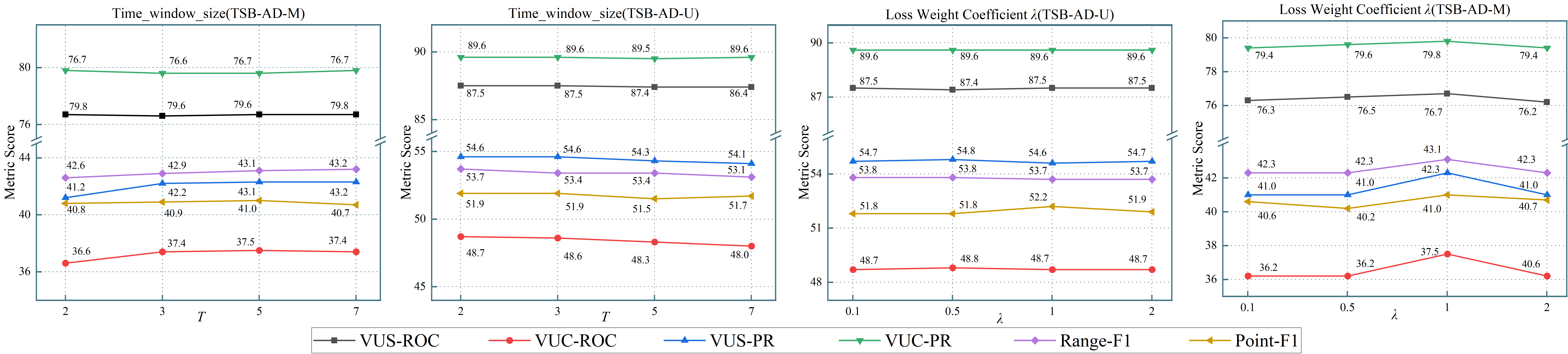}
	\caption{Sensitivity analysis of the performance of univariate and multivariate time-series anomaly detection with respect to window length $T$. All results are presented as percentages (\%).}
	\label{FIG:5}
\end{figure*}

The temporal sequence restoration pretext task shuffles consecutive $T$-patch windows and reconstructs their original order via self-supervised learning. Comparative results with $T\in\{2, 3, 5, 7\}$ are illustrated in Fig.~\ref{FIG:5}. The model achieves stable performance across different window sizes, with no significant drop in accuracy at $T = 7$ compared with $T = 3$ and $T = 5$, validating the strong robustness of the sequence ordering loss to variations in window size. Nevertheless, an excessively large $T$ increases the input dimension of the sequence reconstruction head and raises computational complexity, thereby increasing training and inference latency. To balance detection accuracy and computational efficiency, $T = 2$ and $T = 5$ are adopted for univariate and multivariate tasks, respectively.

The sorting loss weight $\lambda$ balances the optimization objective between the self-supervised pretext task and the primary anomaly detection task. Sensitivity experiments with $\lambda \in \{0.1, 0.5, 1, 2\}$ demonstrate that $\lambda = 1$ yields the optimal overall performance for both univariate and multivariate detection. Slight performance fluctuations are observed under alternative $\lambda$ values, verifying the model’s low sensitivity to this hyperparameter. Such stability effectively reduces parameter tuning costs for industrial deployment. Accordingly, $\lambda = 1$ is set as the default configuration.

To investigate the optimization effect of multiscale kernel configuration on patch representation learning, this study fixes the encoder depth and channel dimension. It conducts comparative experiments with four groups of convolutional kernel combinations. The default kernel setting of PaAno+ is (3, 7, 15), and the comparison groups include (3, 5, 9), (5, 9, 19), and (7, 15, 25). Detailed quantitative results are reported in Table~\ref{tab:6}.

\begin{table*}[htbp]
	\centering
	\caption{Hyperparameter sensitivity analysis for different combinations of convolutional kernels. The optimal and suboptimal values for each metric are indicated in bold and underlined, respectively. All experimental results are presented as percentages (\%).}
	\label{tab:6}
	\small
	\setlength{\tabcolsep}{8pt}   
	\begin{tabular}{llcccccc}
		\toprule
		Dataset & Ablation Variant & VUS-PR & VUS-ROC & Range-F1 & AUC-PR & AUC-ROC & Point-F1 \\
		\midrule
		\multirow{4}{*}{TSB-AD-U}
		&w/o kernel (3,5,9) &53.0&89.4&51.2&46.9&87.2&51.8\\
		&w/o kernel (5,9,19) &53.0&89.3&51.1&47.1&87.2&52.7\\
		&w/o kernel (7,15,25) &53.3&88.9&50.2&47.0&86.8&52.3\\
		&PaAno+ (3,7,15) &\textbf{54.6}&\textbf{89.6}&\textbf{51.9}&\textbf{48.7}&\textbf{87.5}&\textbf{53.7}\\
		\midrule
		\multirow{4}{*}{TSB-AD-M}
		&w/o kernel (3,5,9) &41.6&79.4&40.7&36.6&76.2&42.5\\
		&w/o kernel (5,9,19) &40.1&79.2&39.0&35.3&76.1&40.7\\
		&w/o kernel (7,15,25) &40.5&79.6&39.9&35.8&76.4&41.1\\
		&PaAno+ (3,7,15) &\textbf{42.3}&\textbf{79.8}&\textbf{41.0}&\textbf{37.5}&\textbf{76.7}&\textbf{43.1}\\
		\bottomrule
	\end{tabular}
\end{table*}

On the TSB-AD-U dataset, the proposed (3, 7, 15) configuration achieves optimal or suboptimal results across all six evaluation metrics, with VUS-PR of 54.0\%, VUS-ROC of 89.0\%, Range-F1 of 51.7\%, AUC-PR of 48.0\%, AUC-ROC of 86.2\%, and Point-F1 of 52.9\%. Compared with the second-best (3, 5, 9) setting, the proposed configuration improves VUS-PR by 1.6\%, Range-F1 by 1.6\%, AUC-PR by 2.0\%, and Point-F1 by 1.2\%, respectively. This indicates that kernel combinations with medium receptive-field spans are better suited to capturing local anomalous patterns in univariate time series. The small-scale kernel combination (3, 5, 9) maintains competitive performance, demonstrating the superiority of fine-grained kernels for detecting abrupt local anomalies. In contrast, the large-scale combination (7, 15, 25) causes significant performance degradation, as excessive receptive fields smooth subtle short-term temporal variations and impair the model's ability to characterize anomalies in a fine-grained manner.

On the TSB-AD-M dataset, the (3, 7, 15) kernel configuration outperforms all baseline variants across all metrics, achieving VUS-PR of 42.3\%, VUS-ROC of 79.8\%, Range-F1 of 41.0\%, AUC-PR of 37.5\%, AUC-ROC of 76.7\%, and Point-F1 of 43.1\%. Although the performance gaps among different kernel combinations narrow in multivariate scenarios, the (5, 9, 19) and (7, 15, 25) combinations still exhibit VUS-PR that are 2.2\% and 1.8\% lower than the proposed setting, respectively. This phenomenon is attributed to the inherent cross-variable coupling in multivariate time series, which weakens the model's sensitivity to variations in individual kernel scales.

By combining univariate and multivariate experimental results, the (3, 7, 15) kernel combination achieves an optimal balance between multiscale feature learning. The small kernel of 3 captures fine-grained short-term anomalies, the medium kernel of 7 adapts to conventional periodic temporal patterns, and the large kernel of 15 excavates long-range contextual dependencies. This hierarchical multiscale design enables the encoder to integrate coarse and fine temporal features, providing robust representations for both self-supervised pretext-task optimization and downstream anomaly detection.

In conclusion, PaAno+ achieves stable, optimal performance across a wide range of hyperparameters and requires minimal parameter tuning. Its outstanding robustness and practicality will satisfy the deployment requirements of real-world industrial anomaly detection tasks.

\section{Conclusion}
\label{sec5}
This study proposes an enhanced, lightweight anomaly detection framework, namely PaAno+, to address the inherent limitations of the original PaAno model in receptive-field adaptive matching, cross-variable dependency modeling, and pretext-task supervision. The proposed framework integrates three core designs: a multiscale temporal encoder, a cross-variable fusion attention module, and a sequence reconstruction-based self-supervised pretext task. By embedding parallel multiscale convolutional branches with cross-scale adaptive attention fusion, variable-dimensional multi-head self-attention, and window-level temporal-order restoration learning, PaAno+ substantially strengthens the representation and modeling capabilities for complex anomalous patterns in both univariate and multivariate time series.

Extensive evaluations on the TSB-AD benchmark validate the superiority of PaAno+. In univariate anomaly detection tasks, PaAno+ achieves a VUS-PR of 0.54, demonstrating substantial performance improvements over vanilla PaAno and other baselines. For multivariate detection tasks, PaAno+ achieves VUS-ROC and AUC-ROC scores of 0.80 and 0.77, respectively, with consistent performance gains over the original model. Ablation studies verify the individual efficacy and synergistic effects of all core modules. With only 1.1M and 1.5M parameters for univariate and multivariate tasks, PaAno+ exhibits a much more compact architecture than prevailing Transformer-based detectors. Moreover, hyperparameter sensitivity experiments demonstrate the superior performance stability and robustness of PaAno+ under diverse parameter settings.

Future research directions will focus on lightweight network optimization, adaptive receptive field tuning, and the construction of a universal contrastive learning framework. In addition, further validations on diverse real-world industrial datasets will be conducted to improve the generalization and practical adaptability of the proposed method.

%
%
%

\bibliographystyle{unsrtnat}

\bibliography{cas-refs}

\end{document}